\documentclass{article}

\usepackage{microtype}
\usepackage{graphicx}
\usepackage{subfigure}
\usepackage{booktabs} 

\usepackage{hyperref}

\usepackage[accepted]{icml2025}

\usepackage{amsmath}
\usepackage{amssymb}
\usepackage{mathtools}
\usepackage{amsthm}

\usepackage[capitalize,noabbrev]{cleveref}

\theoremstyle{plain}

\theoremstyle{definition}

\theoremstyle{remark}

\usepackage[textsize=tiny]{todonotes}
\usepackage[normalem]{ulem}
\useunder{\uline}{\ul}{}

\icmltitlerunning{LLM for Extracting Contract Information}

\begin{document}

\twocolumn[
\icmltitle{Large Language Model for Extracting Complex Contract Information in Industrial Scenes}



\icmlsetsymbol{equal}{*}

\begin{icmlauthorlist}
\icmlauthor{Yunyang Cao}{equal,tongji}
\icmlauthor{Yanjun Li}{equal,ecnu}
\icmlauthor{Silong Dai \textsuperscript{\textdagger}}{ecnu}
\end{icmlauthorlist}

\icmlaffiliation{tongji}{Shanghai Research Institute for Intelligent Autonomous Systems, Tongji University, Shanghai, China}
\icmlaffiliation{ecnu}{School of Computer Science and Technology, East China Normal University, Shanghai, China}

\icmlcorrespondingauthor{Silong Dai}{sldai@stu.ecnu.edu.cn}

\icmlkeywords{Large Language Model, Data Labeling, Information Extraction, Data Augmentation}

\vskip 0.3in
]



\printAffiliationsAndNotice{\icmlEqualContribution} 

\begin{abstract}
This paper proposes a high-quality dataset construction method for complex contract information extraction tasks in industrial scenarios and fine-tunes a large language model based on this dataset. Firstly, cluster analysis is performed on industrial contract texts, and GPT-4 and GPT-3.5 are used to extract key information from the original contract data, obtaining high-quality data annotations. Secondly, data augmentation is achieved by constructing new texts, and GPT-3.5 generates unstructured contract texts from randomly combined keywords, improving model robustness. Finally, the large language model is fine-tuned based on the high-quality dataset. Experimental results show that the model achieves excellent overall performance while ensuring high field recall and precision and considering parsing efficiency. LoRA, data balancing, and data augmentation effectively enhance model accuracy and robustness. The proposed method provides a novel and efficient solution for industrial contract information extraction tasks.
\end{abstract}

\section{Introduction}
Contracts play a central role in commercial transactions by clearly defining the rights and obligations of both parties and providing legal protection. With the deepening of digital transformation, the importance of digital management and analysis of contract texts has become increasingly prominent. However, traditional contract management faces numerous challenges, among which the diversity of contract formats and the inefficiency of manual review are particularly critical. Contract information extraction provides a rich, structured data source for building knowledge graphs; in turn, knowledge graphs visually represent the complex relationships between data sources, enabling deeper understanding and application of contract information. This interconnection makes contract management more intelligent, automated, and efficient. In industrial domains, contracts often involve more complex fulfillment requirements, delivery locations, settlement details, material specifications, and additional clauses , covering the entire supply chain.

In recent years, advancements in artificial intelligence, especially in natural language processing (NLP), have provided new approaches for the automated processing of contract texts. Although large language models (LLMs) \cite{ref5,ref6,ref7,ref8} have demonstrated powerful capabilities in downstream tasks such as natural language inference and machine translation \cite{ref9}, their performance in sequence labeling tasks—such as named entity recognition—still lags behind common baselines \cite{ref10}. Moreover, compared to other contexts, industrial contracts are uniquely complex and domain-specific. They cover multiple dimensions, including technical specifications, quality standards, and legal clauses, and frequently involve industry-specific terminology and intricate logical relationships.

Key information extraction from industrial contract texts faces three major challenges:
(1) Text processing capabilities: The complexity of industrial contracts requires models with strong text comprehension abilities to handle specialized vocabulary, lengthy texts, and multi-section structures.
(2) Accuracy and efficiency: Current information extraction in industrial contracts relies heavily on manual review, making it difficult to balance accuracy with efficiency.
(3) Generalization and adaptability: Due to limited data sources, industrial contract texts tend to be uniform in structure with strongly interrelated information.

To address these challenges, this paper proposes an efficient automated data annotation method to construct a large-scale, high-quality dataset of industrial contract texts and significantly enhances the performance of large language models in contract information extraction tasks through fine-tuning. The key \textbf{contributions} of this paper are as follows:

\begin{itemize}
\item For Chinese industrial contracts, an efficient and low-cost automated data annotation method is proposed, using cluster analysis and stratified sampling to construct a high-quality labeled dataset.

\item The dataset is augmented using GPT-4, improving the robustness and generalization ability of the model in downstream tasks.

\item Extensive experiments demonstrate that the proposed method achieves excellent overall performance by ensuring high field recall and precision while maintaining efficient parsing.
\end{itemize}

\section{Related Work}
This section reviews the current state of research from three perspectives: large language models, data annotation methods based on large language models, and information extraction methods based on large language models.

\subsection{Large Language Models} 
Language models \cite{ref11,ref12,ref13} are computational models capable of understanding and generating human language. They can predict the probability of word sequences or generate new text based on a given input. One of the most common types is the n-gram model \cite{ref14}, which estimates the probability of a word based on its context.

Large Language Models (LLMs) are advanced language models with a massive number of parameters and strong learning capabilities. The core components of many LLMs, such as GPT-3 \cite{ref15}, InstructGPT \cite{ref16}, and Qwen \cite{ref17}, are based on the self-attention mechanism in Transformers \cite{ref18}. The Transformer architecture adopts a pre-training and fine-tuning paradigm \cite{ref19,ref20}, allowing effective transfer learning for specific tasks. During fine-tuning, LLMs require sufficient labeled data for training. If training is to be conducted in a specific domain or on a specialized task, it may be necessary to manually process and annotate the data.

\subsection{Data Annotation Methods Based on LLMs}
Data annotation is a critical step in machine learning, and fine-tuning large language models also depends on sufficient labeled data.

Due to the complexity and diversity of data, annotation poses challenges for machine learning models. LLMs, with their strong semantic understanding and text generation capabilities, can automatically perform annotation tasks, ensure data consistency, and adapt to specific domains through fine-tuning or prompting. This significantly improves the effectiveness and accuracy of data labeling and helps address many difficulties inherent in traditional methods \cite{ref21}. ZEROGEN \cite{ref22} demonstrates the practicality of zero-shot prompting with large models. It uses prompts such as “A movie review with a positive sentiment is:” to guide the model in generating text $x$ aligned with label $y$.

\subsection{Information Extraction Methods Based on LLMs}
Information Extraction (IE) refers to the process of extracting specific events or factual information from natural language texts, enabling automatic classification, extraction, and reconstruction of vast amounts of content. This information typically includes entities, relationships, and events. IE mainly comprises three subtasks: Named Entity Recognition (NER) \cite{ref23}, Relation Extraction (RE) \cite{ref24}, and Event Extraction (EE) \cite{ref25}. Common methods include TF-IDF \cite{ref26}, TextRank \cite{ref27}, SnowNLP \cite{ref28}, KeyBERT \cite{ref29}, and UIE \cite{ref30}.

Information extraction can be formulated as a sequence prediction problem. Given an input sequence—such as a sentence or a document—consisting of $n$ tokens $\mathbf{X}=[x_1,\cdots,x_n]$, along with a prompt $P$ and a target sequence 
$\mathbf{Y}=[y_1,\cdots,y_m]$, the objective is to maximize the autoregressive conditional probability:

\begin{equation}
p_{\theta}(\mathbf{Y}|\mathbf{X},P)=\prod_{i=1}^{m}p_{\theta}(y_i|\mathbf{X},P,y_{<i})
\end{equation}
Here, $\theta$ denotes the parameters of the large language model, which may either be frozen or trainable. Several studies \cite{ref31,ref32,ref33} have enhanced the model’s understanding of specific tasks by adding extra prompts or instructions $P$ to the input text $\mathbf{X}$.

\section{Method}
Industrial contract texts are structurally complex and highly diverse, making it difficult to directly generate high-quality annotated data using large language models. To address this challenge, this paper first classifies the texts and then annotates them separately. The annotations from different categories are then aggregated. The approach involves three main components: prompt construction, automated text annotation, and data augmentation. Specifically, we first construct data samples and then generate annotations in batches. To prevent the large language model from relying too heavily on inherent patterns in industrial contract texts, we perform data augmentation by randomly combining keywords from contracts. This forces the model to focus more on the specific content of the clauses. Our contract information extraction method provides rich and structured data sources, integrates domain knowledge from the industrial field, and enables deep understanding and management of contract information.

\subsection{Prompt Construction for Annotation}
In this work, we use GPT-4 to construct data samples and GPT-3.5 to perform batch annotation on the original dataset $\mathbf{D}$. The original industrial contract texts $x$ in $\mathbf{D}$ come from various business backgrounds. Using a single prompt $P$ and data sample to annotate all texts often leads to incomplete extraction of key information. To solve this problem, we first perform cluster analysis on the original dataset $\mathbf{D}$ and use GPT-4 to extract information from the samples $c_i$ closest to the cluster centers $\mu_i$ in each cluster. The extracted result is taken as the annotation $y_{c_i}$. Each annotated pair $\{c_i, y_{c_i}\}$, along with the prompt $P$, is then used as input to GPT-3.5 for batch annotation.

Specifically, we first encode the texts using TF-IDF \cite{ref34} and then apply the K-means algorithm to cluster the encoded texts. To achieve the optimal clustering result, we consider the number of clusters $K \in [2, 20]$, where $K$ is an integer, and use the Within-Cluster Sum of Squares (WCSS) as the clustering score to evaluate the results. The final number of clusters is chosen based on the highest clustering score. The WCSS is defined as follows:

\begin{equation}
    WCSS = \sum_{i=1}^{k}{\sum_{x\in\mathbf{S}_i}||x-\mu_i||^2}
\end{equation}
Where $k$ is the number of clusters, $\mathbf{S}_i$ is the set of data points in the $i$‑th cluster, $\mu_i$ is the centroid of the $i$‑th cluster, $x$ is a data point in the cluster, and $||x-\mu_i||$ is the Euclidean distance from $x$ to the cluster center $\mu_i$.

\begin{algorithm}[tb]
   \caption{Clustering Original Data Based on TF-IDF}
   \label{alg:cluster}
\begin{algorithmic}
    \STATE {\bfseries Input:} Original text data $D$, TF-IDF model, Maximum number of clusters $K$, K-means clustering algorithm
    \STATE {\bfseries Output:} Clustered dataset $C$
    \STATE $X = \text{TF-IDF}(D)$           
    \STATE $k = 2$
    \STATE $best_{k} = 0$
    \STATE $best_{wcss} = 0$
    \WHILE{$k \leq K$}
    \STATE $S = \text{K-means}(X, k)$   
    \STATE $sum = 0$
    \FOR{{\bfseries each} $element$ {\bfseries in} $S$}
    \STATE $sum = sum + element$
    \ENDFOR
    \STATE $\mu = sum / length(S)$ 
    \STATE $WCSS = 0$
    \FOR{{\bfseries each} $s$ {\bfseries in} $S$}
    \STATE $WCSS = WCSS + (s - \mu)\times(s - \mu)$ 
    \ENDFOR
    \IF{$best_{wcss} \geq WCSS$}
    \STATE $best_{wcss} = WCSS$
    \STATE $best_{k} = k$
    \ENDIF
    \STATE $k = k + 1$
    \ENDWHILE
    \STATE $C = \text{K-means}(X, best_{k})$
\end{algorithmic}
\end{algorithm}

After obtaining the clustering result $\mathbf{C}$, we extract the same number of samples from each cluster $\mathbf{C}_i$ to form a subset $\mathbf{C}_i^{\prime}$ for annotation, which reduces both annotation cost and model fine-tuning time. We then proceed with automated text annotation.

\subsection{Automated Text Annotation Method} 
Automated text annotation can significantly reduce the cost of data labeling while greatly improving annotation efficiency. It serves as the foundation for the large-scale generation of high-quality fine-tuning datasets.

Specifically, after obtaining the clustering results $\mathbf{C}$ from the original dataset $\mathbf{D}$, GPT-4 is used to construct data samples $\{c_i,y_{c_i}\}$ for each cluster $\mathbf{C}_i$. Using the contextual learning capabilities of large language models, these data samples are incorporated into the prompt $P$ and GPT-3.5 is used to generate annotation results $\mathbf{Y}_i$ for each subset $\mathbf{C}_i^{\prime}$. The detailed procedure is as follows:

\begin{itemize}
    \item Step 1: For each subset $\mathbf{C}_i^{\prime}$, the sample $c_i$ that is closest to the cluster center $\mu_i$ is selected as input, and GPT-4 is used to generate the corresponding annotation result $y_{c_i}$.

    \item Step 2: The pair $\{c_i,y_{c_i}\}$ is used as a data example. With prompt $P$, GPT-3.5 annotates the remaining contract texts $x_i$ in $\mathbf{C}_i^{\prime}$:
    \begin{equation}
        \mathbf{Y}_i=\text{GPT-3.5}(c_i, y_{c_i},P,x_i), \forall x_i\in \mathbf{C}_i^{\prime}
    \end{equation}
    
    \item Step 3: A portion of the annotation results in $\mathbf{Y}_i$ is sampled for manual review to ensure quality.

    \item Step 4: Steps 1 to 3 are repeated until annotations $Y$ have been generated for all subsets $\mathbf{C}^{\prime}$
\end{itemize}

\begin{figure}[ht]
\vskip 0.2in
\begin{center}
\centerline{\includegraphics[width=\columnwidth]{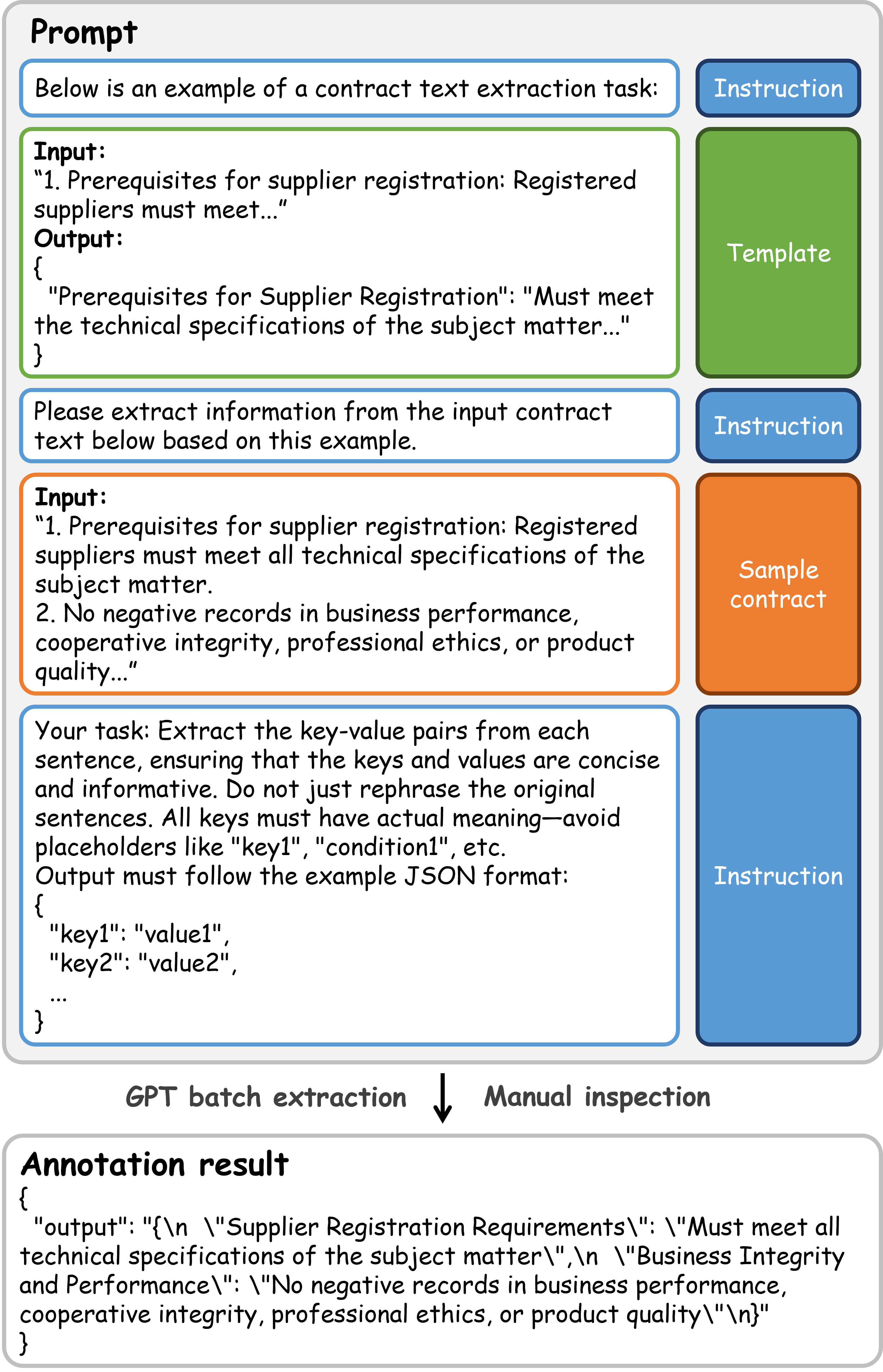}}
\caption{The workflow of the automated text annotation process.}
\label{workflow}
\end{center}
\vskip -0.2in
\end{figure}

This process fully leverages the textual understanding and contextual learning capabilities of large language models. It significantly improves annotation efficiency while maintaining high-quality standards. Through these four steps, we transform the original unlabeled dataset $\mathbf{D}$ into a high-quality labeled dataset $\mathbf{D}^{\prime}=\{\mathbf{C}^{\prime},\mathbf{Y}\}$, which can be directly used for fine-tuning large models. An example of the automated annotation process is shown in \cref{workflow}.

\subsection{Data Augmentation Techniques}
During the construction of the contract text dataset, because annotations are generated independently for each cluster subset $\mathbf{C}_i^{\prime}$, the model may become overly reliant on specific keyword combinations and overlook the actual content of contract clauses. This paper aims to encourage the model to focus more on the detailed content of contract clauses rather than the global distribution of keywords.

Since the model architecture cannot be modified, we propose an innovative data augmentation strategy based on cluster analysis to reduce the model's dependence on fixed keyword combinations. This promotes the model’s attention to the substantive content of the contract clauses, thereby enhancing its ability to identify and process previously unseen contract samples and improving its performance in new scenarios.

Specifically, the proposed data augmentation method involves the following steps:
\begin{itemize}
    \item Step 1: Extract all keywords that appear in the annotated data $\mathbf{Y}$, and construct a global keyword set $\mathbf{K}$.

    \item Step 2: Randomly sample 5–6 keywords from the set $\mathbf{K}$ to form a new keyword combination $\mathbf{k}$.

    \item Step 3: Repeat Step 2 until the predefined number of keyword combinations is obtained, forming a new set of structured annotations $\mathbf{K}^{\prime}$.

    \item Step 4: Use the annotations $\mathbf{K}^{\prime}$ as input and prompt a large language model to generate new unstructured contract texts $\mathbf{X}^{\prime}$, thereby building the augmented dataset $\hat{\mathbf{D}}=\{\mathbf{X}^{\prime},\mathbf{K}^{\prime}\}$.
\end{itemize}

\begin{figure}[ht]
\vskip 0.2in
\begin{center}
\centerline{\includegraphics[width=\columnwidth]{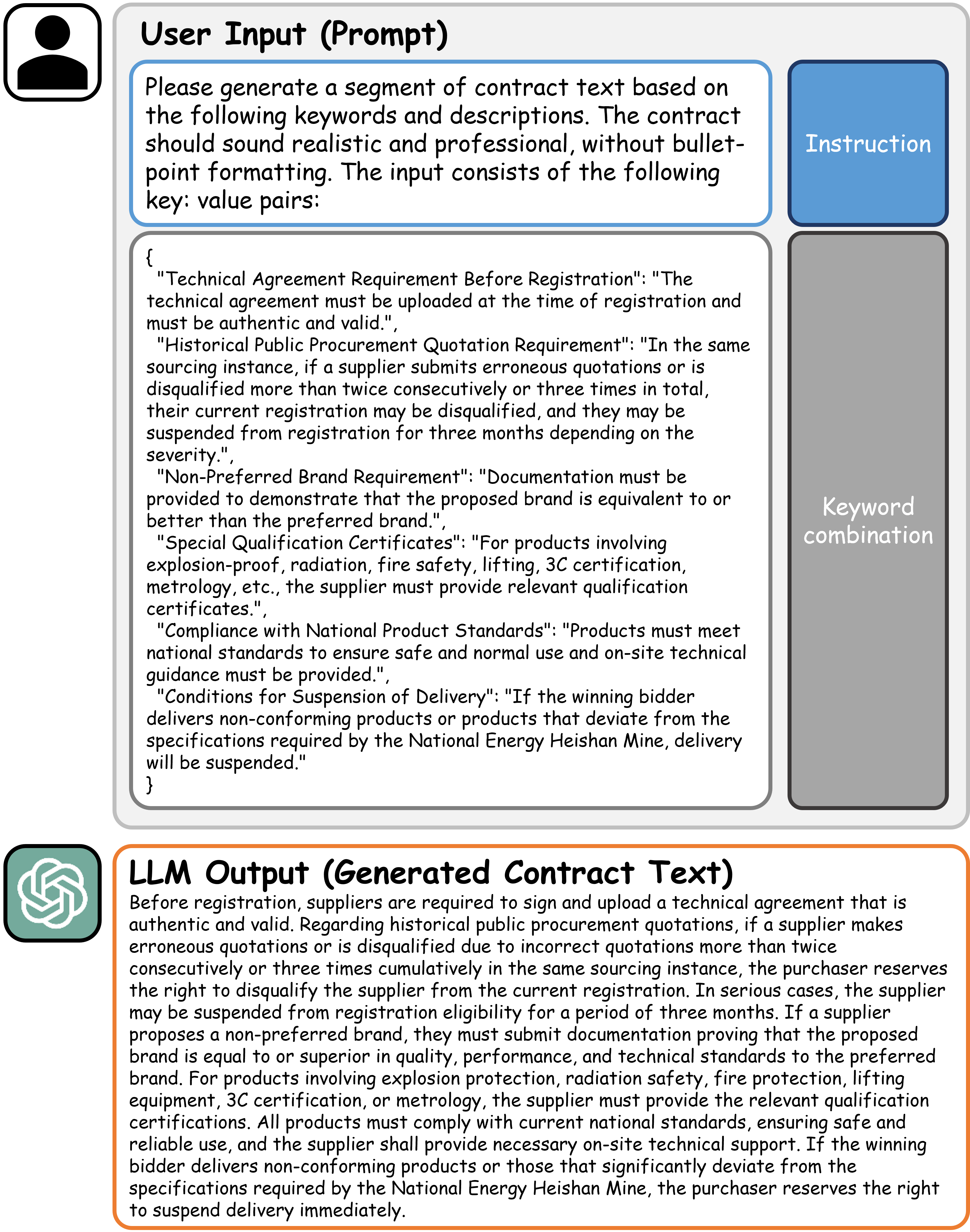}}
\caption{An example of data augmentation.}
\label{generate}
\end{center}
\vskip -0.2in
\end{figure}

This method generates an augmented dataset with altered keyword combination patterns, resulting in entirely new contract texts. It enriches the diversity of training data, helping the model focus more on the actual clauses of the contracts. An example of this data augmentation process is shown in \cref{generate}.

\section{Experimental Results and Analysis}
To validate the effectiveness of the proposed large language model for complex contract information extraction, we conduct experiments on an industrial contract dataset. The objective is to investigate the following three research questions:

RQ1: How effective is the proposed large language model for information extraction?

RQ2: Does LoRA fine-tuning improve the extraction accuracy of the large language model?

RQ3: What is the impact of data balancing and data augmentation on the model’s extraction accuracy and robustness?

\subsection{Experimental Data}
The dataset used in this study is the Intelligent Contract Information Extraction Dataset \cite{dataset}. It is provided by the Obei Industrial Supply Chain Ecosystem Platform and includes contracts for bidding, procurement, sales, transportation, and services. The platform divides the dataset into training and testing sets. \textbf{All contracts in the dataset are written in Chinese}. Each contract can be divided into two parts based on content: qualification text and requirement text. This study counts the number of samples and the number of characters (i.e., text length) for each part. Basic statistics of the dataset are shown in \cref{tab:statistics}, where Q stands for Qualification and R stands for Requirement.

\begin{table}[htb]
\caption{Dataset statistics.}
\label{tab:statistics}
\vskip 0.15in
\begin{center}
\begin{small}
\begin{sc}
\begin{tabular}{cccc}
\toprule
Set Type & Text Type & \#Samples & Text Length\\
\midrule
Training & Q & 10,000 & [51, 1000] \\
 & R & 10,000 & [53, 780] \\
Test & Q & 174 & [57, 264] \\
 & R & 174 & [65, 503] \\
\bottomrule
\end{tabular}
\end{sc}
\end{small}
\end{center}
\vskip -0.1in
\end{table}

\begin{figure}[ht]
\vskip 0.2in
\begin{center}
\centerline{\includegraphics[width=\columnwidth]{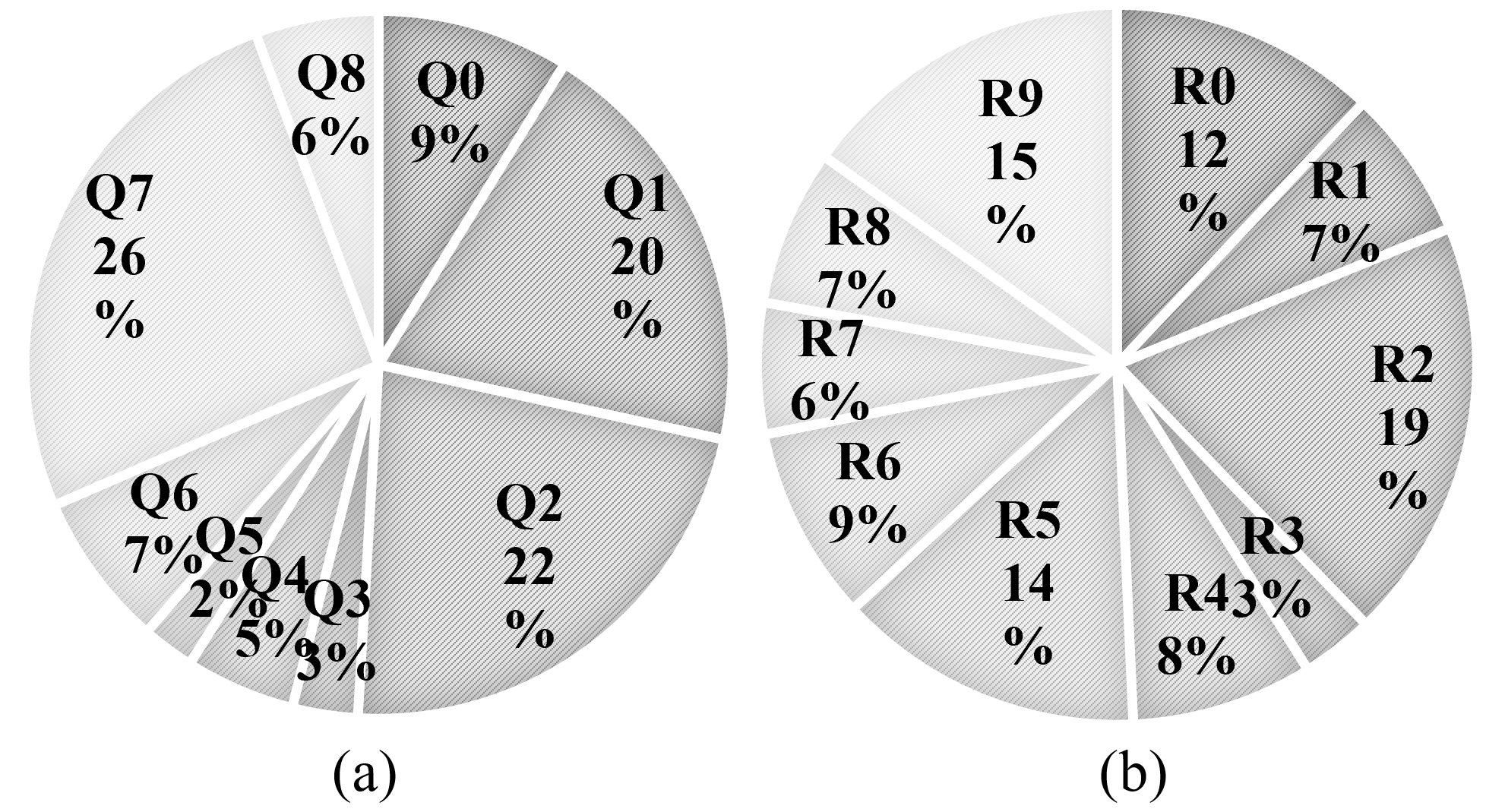}}
\caption{Category distribution ratio of qualification text and requirement text.}
\label{pie}
\end{center}
\vskip -0.2in
\end{figure}

According to the method described in Section 3.1, we perform clustering analysis on the dataset. Specifically, we apply K-Means clustering separately to the qualification texts and requirement texts, dividing the qualification texts into 9 clusters (denoted as Q0, Q1, ..., Q8) and the requirement texts into 10 clusters (denoted as R0, R1, ..., R9). The distribution proportions of the qualification text clusters and requirement text clusters are shown in \cref{pie}(a) and (b), respectively.

From \cref{pie}, it can be observed that both qualification and requirement texts suffer from class imbalance. Therefore, stratified sampling is necessary to balance the sample distribution more effectively.

The Intelligent Contract Information Extraction Dataset serves as the experimental foundation for our method. The subsequent sections will analyze model performance based on evaluation metrics.

\subsection{Evaluation Metrics and Base Models}
This study evaluates model performance from three key aspects: field extraction accuracy, field recall, and efficiency. Specifically:
\begin{itemize}
    \item Field extraction accuracy assesses whether the structured output keys match the true keys.
    \item Field recall evaluates whether the structured output fully matches the gold-standard structured text.
    \item Efficiency measures the model's speed in parsing contract texts.
\end{itemize}
The specific metrics used for each aspect are described below.

\textbf{Field Extraction Accuracy.} Field accuracy measures the hit rate of key fields predicted by the model. The mathematical expressions are as follows:
\begin{equation}
    acc_i=\frac{||\mathbf{Key}_{i}^{match}||}{||\mathbf{Key}_{i}^{label}||},
\end{equation}
\begin{equation}
    acc=\frac{1}{n}\sum_{i=1}^n acc_i,
\end{equation}
where $\mathbf{Key}_{i}^{match}$ is the set of fields correctly extracted by the model for the $i$-th sample. $\mathbf{Key}_{i}^{label}$ is the set of true fields for the $i$-th sample. $n$ is the total number of contract samples. $||\cdot||$ denotes the number of elements in a set. $acc_i$ is the field extraction accuracy for the $i$-th sample. $acc$ is the average field extraction accuracy across all samples. A predicted field is considered correct if its edit distance from the labeled field is no greater than 3. This metric is calculated separately for qualification text, requirement text, and all texts combined.

\textbf{Field Recall.} Field recall measures how well the structured content covers the fields commonly present in contracts. To approximate this metric, we adopt the ROUGE (Recall-Oriented Understudy for Gisting Evaluation) scores \cite{ref35}. We use ROUGE-1, ROUGE-2, and ROUGE-L to evaluate the model. ROUGE-N measures the n-gram overlap between the generated summary and the reference summary, with $N$ denoting the length of the n-grams. ROUGE-L measures the longest common subsequence (LCS) overlap. The formula for ROUGE-N and ROUGE-L are:
\begin{equation}
    \text{ROUGE-N}=\frac{\sum_{\text{references}}\sum_{\text{n-gram}\in\text{candidate}} N_{\text{match}}(\text{n-gram})}{\sum_{\text{references}}\sum_{\text{n-gram}\in\text{references}} N(\text{n-gram})}
\end{equation}
\begin{equation}
    \text{ROUGE-L}=\frac{\sum_{\text{references}} LCS(\text{candidate, references})}{\sum_{\text{references}} Length(\text{references})}
\end{equation}
where $N_{\text{match}}(\text{n-gram})$ is the count of overlapping n-grams between candidate and reference. $LCS(\cdot)$ is the length of the longest common subsequence. $Length(\cdot)$ denotes the total length of the reference summary.

\textbf{Efficiency.} Efficiency refers to the model’s contract parsing speed, which is the reciprocal of the average time taken to parse a contract. Therefore, we use \textbf{contracts parsed per second} as the evaluation metric.

\textbf{Base Models.}
Three open-source Chinese large language models are used as base models for comparison: \textbf{GLM-4-9B (GLM4)} \cite{ref36}, \textbf{Baichuan2-13B-Chat (Baichuan2)} \cite{ref37}, \textbf{Qwen2.5-14B (Qwen2.5)} \cite{ref17}. These models are known for strong performance and have fewer than 20 billion parameters, making them suitable for fine-tuning on a single GPU.

All seven selected evaluation metrics are "the higher, the better." Detailed experimental results will be discussed in Sections 4.4 to 4.6.

\subsection{Experimental Environment and Basic Parameters} 
The experiments in this study are conducted on a single NVIDIA RTX A6000 GPU (48GB). The model fine-tuning approach adopted is LoRA (Low-Rank Adaptation). The number of training epochs and the LoRA rank are treated as tunable hyperparameters, and optimal parameter analysis will be discussed in Section 4.5. 
In the performance evaluation in Section 4.4, we compare the performance of the three base models, GLM4, Baichuan2, and Qwen2.5, under the same hyperparameter settings. Additionally, four traditional information extraction methods are used as baselines for comparison: TF-IDF, TextRank, SnowNLP, KeyBERT. These methods are used to extract keywords, and the UIE (Universal Information Extraction) framework is applied to extract key information based on those keywords.
In the ablation study in Section 4.6, we evaluate the impact of data balancing and data augmentation on model performance under the same hyperparameter settings. During data augmentation, 10\% new samples are generated.

With the above, all experimental preparations have been completed. The following sections will present the experiments and analyze the results.

\subsection{Performance Evaluation}
\begin{table*}[t]
\caption{Main results.}
\label{tab:performance}
\vskip 0.15in
\begin{center}
\begin{small}
\begin{sc}
\begin{tabular}{lccccccc}
\toprule
              & Q acc. & R acc. & Total acc. & ROUGE-1 & ROUGE-2 & ROUGE-L & Efficiency \\
\midrule
TF-IDF+UIE    & 0.1505 & 0.1057 & 0.1281 & 0.0000  & 0.0000  & 0.0000  & 0.31    \\
TextRank+UIE  & 0.0259 & 0.1282 & 0.0770 & 0.0012  & 0.0000  & 0.0012  & 0.31    \\
SnowNLP+UIE   & 0.1621 & 0.2391 & 0.2006 & 0.0003  & 0.0000  & 0.0003  & 0.31    \\
KeyBERT+UIE   & 0.0396 & 0.0523 & 0.0460 & 0.0012  & 0.0000  & 0.0012  & 0.30    \\
GLM4          & 0.1709 & 0.6136 & 0.3819 & 0.6173  & 0.4757  & 0.5697  & 0.55    \\
Baichuan2     & 0.1745          & 0.7156          & 0.4240          & 0.6202          & 0.4772          & 0.5633          & \textbf{0.81} \\
Qwen2.5       & \textbf{0.2495} & {\ul 0.7161}    & {\ul 0.4665}    & {\ul 0.6290}    & {\ul 0.4929}    & {\ul 0.5712}    & {\ul 0.57}    \\
Qwen2.5-Final & {\ul 0.2431}    & \textbf{0.7266} & \textbf{0.4724} & \textbf{0.6296} & \textbf{0.5008} & \textbf{0.5791} & 0.48 \\  
\bottomrule
\end{tabular}
\end{sc}
\end{small}
\end{center}
\vskip -0.1in
\end{table*}

In the performance evaluation, eight different methods are compared. Based on initial hyperparameter tuning, Qwen2.5 demonstrates the best performance; therefore, further fine-tuning is applied to Qwen2.5, and the resulting model is referred to as Qwen2.5-Final.

\cref{tab:performance} presents the performance of all eight methods. Among them, GLM4, Baichuan2, and Qwen2.5 are trained under the same hyperparameter settings: one training epoch and LoRA rank of 8. Qwen2.5-Final is trained with the optimal hyperparameters. In \cref{tab:performance}, Text Q refers to qualification text, and Text R refers to requirement text. For each metric, the highest value is shown in bold, and the second highest value is underlined.

\begin{figure}[ht]
\vskip 0.2in
\begin{center}
\centerline{\includegraphics[width=\columnwidth]{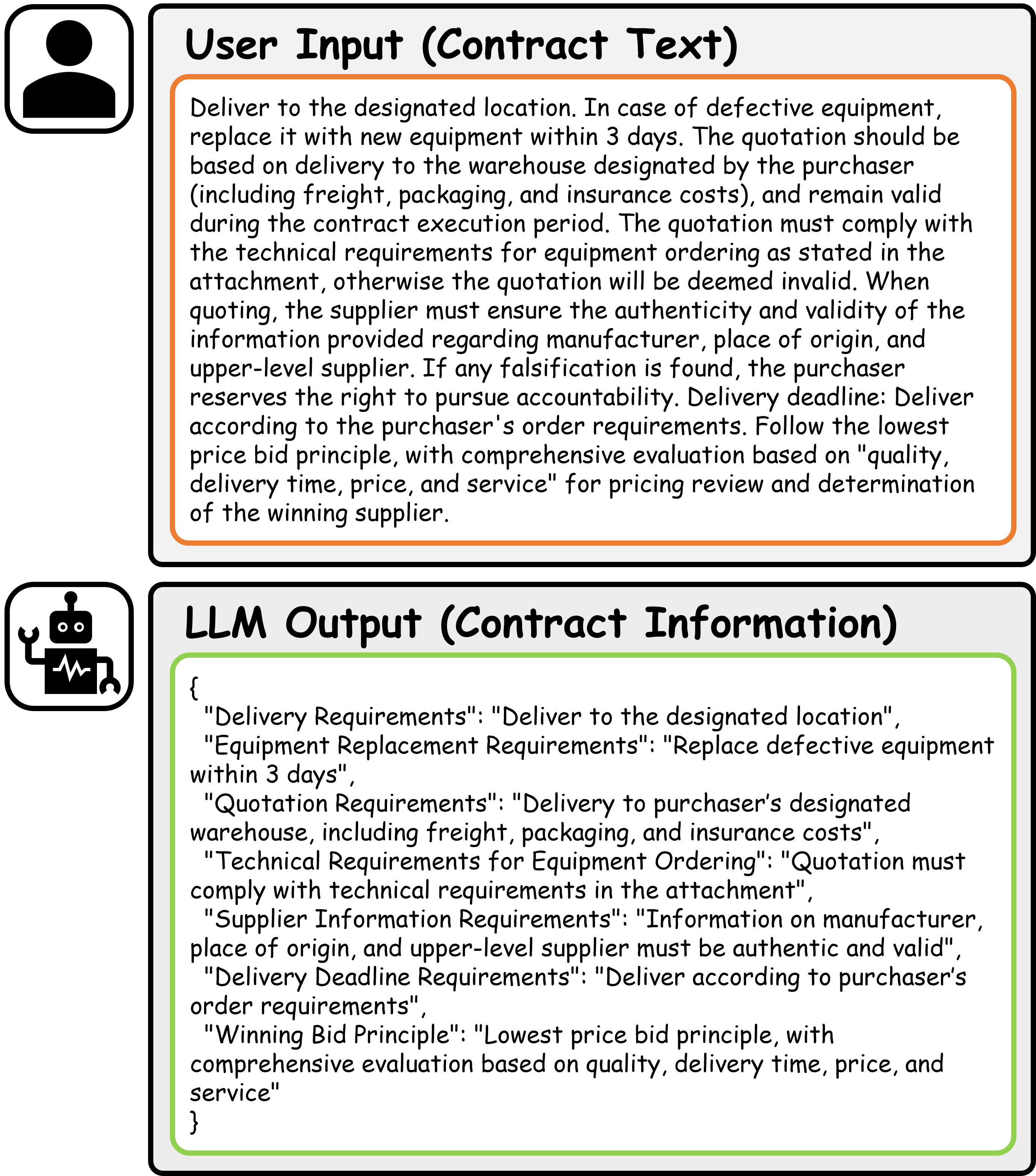}}
\caption{An example of model extracting information from contract text.}
\label{result}
\end{center}
\vskip -0.2in
\end{figure}

In addition, we performed a qualitative analysis of the model outputs. \cref{result} shows an example of information extraction from contract texts by the model.

Based on the results in the performance test and the extraction example, we can address Research Question 1 (RQ1): The large language models demonstrate extraction performance that meets real-world business requirements. Specifically, the large language models significantly outperform traditional extraction methods across all metrics.

Among the fine-tuned large models, Qwen2.5 achieves the highest field extraction accuracy, highest recall rate, and the second-fastest processing efficiency. Therefore, Qwen2.5 is selected as the base model.

After applying optimal hyperparameter settings, the final model Qwen2.5-Final exhibits substantial improvements in field recall rate and ROUGE scores. In particular, the model successfully extracts key information such as: “Equipment replacement requirements” from performance clauses, “Delivery requirements” from delivery location clauses, “Quotation requirements” from settlement sections, “Technical requirements for equipment ordering” from material information, and “Bidding principles” from additional terms.

In conclusion, the proposed method offers fast automated extraction, significantly higher accuracy than traditional non-LLM methods, and domain-specific knowledge relevant to industrial contracts, meeting the practical needs of business applications.

\subsection{Optimal Parameter Analysis}
To investigate the impact of hyperparameters on model training, this study analyzes two key hyperparameters of LoRA fine-tuning: the rank and the number of training epochs, aiming to determine the optimal parameter combination for fine-tuning the model.

The LoRA rank is a critical hyperparameter that determines the number of trainable parameters during fine-tuning. As shown in \cref{tab:rank}, we compare the performance of Qwen2.5 under different rank settings with the number of training epochs fixed at 1. The results indicate that the model achieves the best performance when the LoRA rank is 12.

\begin{table}[ht]
\caption{Model results under three LoRA rank parameter settings.}
\label{tab:rank}
\vskip 0.15in
\begin{center}
\begin{small}
\begin{sc}
\begin{tabular}{cccc}
\toprule
LoRA rank & ROUGE-1 & ROUGE-2 & ROUGE-L \\ \midrule
4         & 0.6283  & 0.4952  & 0.5734  \\
8         & 0.6292  & 0.4929  & 0.5712  \\
12        & 0.6296  & 0.5008  & 0.5791  \\ \bottomrule
\end{tabular}
\end{sc}
\end{small}
\end{center}
\vskip -0.1in
\end{table}

Additionally, we further evaluate the effect of different training epoch counts on model performance, keeping the LoRA rank fixed at 12. The results, summarized in \cref{tab:epoch}, show that the model achieved the best overall performance based on the three ROUGE metrics, when the number of training epochs is 1.

\begin{table}[ht]
\caption{Model results under five epochs parameter settings.}
\label{tab:epoch}
\vskip 0.15in
\begin{center}
\begin{small}
\begin{sc}
\begin{tabular}{cccc}
\toprule
epochs & ROUGE-1 & ROUGE-2 & ROUGE-L \\ \midrule
1      & 0.6296  & 0.5008  & 0.5791  \\
2      & 0.6233  & 0.4852  & 0.5598  \\
3      & 0.6173  & 0.4753  & 0.5632  \\
4      & 0.6160  & 0.4717  & 0.5609  \\
5      & 0.6228  & 0.4968  & 0.5876  \\ \bottomrule
\end{tabular}
\end{sc}
\end{small}
\end{center}
\vskip -0.1in
\end{table}

Therefore, we select the model trained with a LoRA rank of 12 and 1 training epoch as the final model configuration for subsequent experiments.

\subsection{Ablation Study}
To address Research Question RQ2, we design ablation experiments to verify the effectiveness of fine-tuning in improving model performance. Using LoRA for fine-tuning, we compare model performance before and after fine-tuning. The results are presented in \cref{tab:ft}.

\begin{table*}[tb]
\caption{Performance comparison before and after fine-tuning. R-1, R-2, R-L are the abbreviations of ROUGE-1, ROUGE-2, and ROUGE-L respectively.}
\label{tab:ft}
\vskip 0.15in
\begin{center}
\begin{small}
\begin{sc}
\begin{tabular}{lrrrrrrrrr}
\toprule
          & \multicolumn{3}{c}{After Fine-tuning}                                                   & \multicolumn{3}{c}{Before Fine-tuning}                                                  & \multicolumn{3}{c}{Improvement Rate}                                                    \\
          & \multicolumn{1}{c}{R-1} & \multicolumn{1}{c}{R-2} & \multicolumn{1}{c}{R-L} & \multicolumn{1}{c}{R-1} & \multicolumn{1}{c}{R-2} & \multicolumn{1}{c}{R-L} & \multicolumn{1}{c}{R-1} & \multicolumn{1}{c}{R-2} & \multicolumn{1}{c}{R-L} \\ \midrule
GLM4      & 0.6173                      & 0.4757                      & 0.5697                      & 0.4904                      & 0.2920                      & 0.3347                      & +25.88\%                    & +62.91\%                    & +70.21\%                    \\
Baichuan2 & 0.6202                      & 0.4772                      & 0.5633                      & \textbf{0.5038}             & \textbf{0.3190}             & \textbf{0.3783}             & +23.10\%                    & +49.59\%                    & +48.90\%                    \\
Qwen2.5   & \textbf{0.6292}             & \textbf{0.4929}             & \textbf{0.5712}             & 0.4445                      & 0.2051                      & 0.2755                      & \textbf{+41.55\%}           & \textbf{+140.32\%}          & \textbf{+107.33\%} \\
\bottomrule
\end{tabular}
\end{sc}
\end{small}
\end{center}
\vskip -0.1in
\end{table*}

As shown in \cref{tab:ft}, although Qwen2.5 performs the worst before fine-tuning and Baichuan2 haa the best initial performance, Qwen2.5 shows the most significant improvement after LoRA fine-tuning, achieving the highest ROUGE-1, ROUGE-2, and ROUGE-L scores among all models. This confirms that LoRA fine-tuning significantly enhances the extraction accuracy of large language models.

To address Research Question RQ3, another ablation study is conducted to examine the impact of data balancing and data augmentation on model performance. We evaluate four processing conditions, each trained independently three times. We compute the mean and standard deviation of ROUGE-1 scores to assess both accuracy and robustness. Results are shown in \cref{tab:balance}.

\begin{table}[ht]
\caption{Impact of data balancing and augmentation on ROUGE-1.}
\label{tab:balance}
\vskip 0.15in
\begin{center}
\begin{small}
\begin{sc}
\begin{tabular}{lcc}
\toprule
Processing Type              & Mean    & Std. Dev. \\ \midrule
Imbalanced + No Aug. & 0.6217          & 0.0021            \\
Balanced + No Aug.   & 0.6274          & 0.0017            \\
Imbalanced + Aug.    & 0.6261          & \textbf{0.0009}   \\
Balanced + Aug.      & \textbf{0.6289} & 0.0011            \\ \bottomrule
\end{tabular}
\end{sc}
\end{small}
\end{center}
\vskip -0.1in
\end{table}

From \cref{tab:balance}, we observe that data balancing significantly improves prediction accuracy (higher ROUGE-1 mean). Data augmentation improves both accuracy and model robustness (lower standard deviation). The combined use of balancing and augmentation yields the best performance in both accuracy and robustness. Therefore, the proposed data preprocessing strategies are both feasible and effective, and contribute meaningfully to improving model performance in industrial contract information extraction.

\section{Conclusion}
To address the challenge of extracting information from industrial contract texts, this paper proposes an automated data annotation and augmentation method based on large language models. Leveraging the advanced capabilities of GPT-4 and GPT-3.5, the approach performs clustering and key information extraction on raw contract data to build a high-quality annotated dataset. Moreover, by decoupling the implicit binding relationships among keywords, and having GPT-3.5 annotate randomly recombined keywords, the method enhances the model’s adaptability to varied contract scenarios.

In addition, this research deploys open-source large language models locally, and conducts fine-tuning and optimization specifically for the industrial contract domain. Experimental results demonstrate that the proposed approach achieves high field-level recall and accuracy while maintaining efficient parsing performance. Furthermore, data balancing and augmentation significantly improve both accuracy and robustness, yielding strong overall performance.

The future work can focus on further balancing model performance with computational cost, and exploring lightweight and modular optimization strategies. These directions are essential for promoting the practical deployment of this technology in industrial applications.


\bibliography{references}
\bibliographystyle{icml2025}
\end{document}